\documentclass[a4paper, 10pt, conference]{IEEEtran}
\IEEEoverridecommandlockouts
\usepackage{cite}
\usepackage{amsmath,amssymb,amsfonts}
\usepackage{algorithmic}
\usepackage{graphicx}
\usepackage{multirow}
\usepackage{textcomp}
\usepackage{xcolor}
\def\BibTeX{{\rm B\kern-.05em{\sc i\kern-.025em b}\kern-.08em
    T\kern-.1667em\lower.7ex\hbox{E}\kern-.125emX}}
\newcommand{\etal}{\textit{et al. }}

\begin{document}

\title{AtteSTNet - An attention and subword tokenization based approach for code-switched Hindi-English hate speech detection\\
% \thanks{Equal Contribution by both authors.}
}

\author{\IEEEauthorblockN{Geet Shingi*}
\IEEEauthorblockA{
\textit{Department of Computer Science}\\
University of Southern California \\ California, United States of America \\
geet.shingi@gmail.com}
\and
\IEEEauthorblockN{Vedangi Wagh*}
\IEEEauthorblockA{
\textit{Fu Foundation School of Engineering and Applied Science} \\ Columbia University \\
New York, United States of America \\
vedangikwagh@gmail.com}
}

\maketitle

\begin{abstract}
Recent advancements in technology have led to a boost in social media usage which has ultimately led to large amounts of user-generated data which also includes hateful and offensive speech. The language used in social media is often a combination of English and the native language in the region. In India, Hindi is used predominantly and is often code-switched with English, giving rise to the Hinglish (Hindi+English) language. Various approaches have been made in the past to classify the code-mixed Hinglish hate speech using different machine learning and deep learning-based techniques. However, these techniques make use of recurrence on convolution mechanisms which are computationally expensive and have high memory requirements.  Past techniques also make use of complex data processing making the existing techniques very complex and non-sustainable to change in data. We propose a much simpler approach which is not only at par with these complex networks but also exceeds performance with the use of subword tokenization algorithms like BPE and Unigram along with multi-head attention-based technique giving an accuracy of  \textbf{87.41\% }and F1 score of \textbf{0.851} on standard datasets. Efficient use of BPE and Unigram algorithms help handle the non-conventional Hinglish vocabulary making our technique simple, efficient and sustainable to use in the real world.
\end{abstract}


\section{Introduction}
\par
With easy access to technology, social media has seen a rapid increase in usage over the globe. Every individual has a smartphone and an instant access to social media sites like Facebook, Twitter. These social media networks generate massive amounts of data daily which also contains huge amounts of hate speech. The term "hate speech" can be defined in many ways and its definition changes from person to person but to generalize the definition we can say that any form of speech or writing that denigrates and belittles another person's beliefs, views, or orientation especially based on race, sexual orientation or religion is hate speech.
\par
Now since social media users are around the globe the text data that’s generated also doesn't have any limitation on language. It is largely observed that English with native languages is predominantly used on social media. Focusing on social media users in India the text content that's generated largely has a general trend of containing some English, Hindi, and code-mixed Hindi words and sentences. Hate speech is often discouraging and can have adverse effects on people as it forms a part of cyberbullying. Detecting this type of speech can be useful for identifying users and for imposing strict actions on them. Hate speech detection in a code-mixed language is particularly a challenge due to its nature of having the essence of more than one language.
\par
Previous approaches for Hindi-English code-switched language have used various machine learning and deep learning algorithms. Advanced deep learning-based approaches have also made use of concurrence and recurrence mechanisms \cite{joshi, lal}. However, the use of such mechanisms increases the complexity of the architecture. Further, encoding the Hindi-English texts is also a challenge. Past approaches have either made use of a manually created profanity list \cite{chopra} for Hinglish language or made use of translation \cite{gupta, mathur-2018-trans} to convert the Hindi words to English. Also, the approaches make use of a manually created dictionary for translation of some Hinglish words when the automatic translation fails. However, usage of such approaches in the real world might be difficult due to such complex data processing steps. And in case the model has to be modified due to a change in the data, it would take a lot of time to modify the profanity list and the translation dictionary. Therefore, an approach that is simple, efficient, and sustainable is the need of the hour. 
\par
In this paper, we propose the use of a simple and sustainable model architecture using an attention mechanism along with Byte Pair Encoding (BPE) and Unigram subword tokenization algorithms. Particularly, we make use of multi-head self-attention. An individual text is encoded using BPE as well as Unigram algorithms and the encoded sequences are passed on to their respective Positional Encoding and Attention layers. The outputs obtained are then concatenated and passed on to further layers of the model architecture. The proposed architecture is found to be superior in handling code-switched Hindi-English language hate speech detection and provides optimum results on the standard dataset proposed in \cite{mathur-2018-trans} across various metrics. We also compare our obtained results with previous approaches used by past researchers for the same dataset and show the superior performance of the proposed approach.
\par Our main contributions in the paper could be listed as follows: 
\begin{itemize}
    \item We have made effective use of BPE and Unigram algorithms to handle the non-conventional Hinglish vocabulary without requiring very complex data processing steps like manually creating a profanity list or translating the Hindi words to English in the sentence, unlike past approaches.
    \item We have achieved quality results by using attention despite eliminating the whole recurrence mechanism which is used in most approaches.
    \item We made constructive use of positional encoding in absence of recurrence to provide sequence-related information.
    \item We have been able to illustrate effective retention of maximum information present in a sequence by using the concatenation of BPE and Unigram padded sequences. 
\end{itemize}

The rest of the paper is structured as follows: Section 2 talks about related work in this area while the proposed methodology is explained in section 3. Dataset description and the results obtained are discussed in section 4. Our analysis of the work conducted is presented in section 5 while the paper is concluded in section 6.

\section{Related Work}
\par
Hate speech detection has been an active area of research in the field of natural language processing. Multiple approaches in the past have been tried in the area of hate speech detection, especially after the rise in the use of machine learning algorithms as well as the availability of computational power and data. However, most of the studies in hate speech detection are based on monolingual content, primarily English. Dinakar \etal \cite{dinakar} proposed the use of various features like Tf-Idf, PoS tagging, and label-specific features to detect offensive tweets. Badjatiya \etal \cite{badjatiya} proposed the use of multiple approaches like CNNs, LSTMs, and FastText for hate speech detection. Pitsilis \etal \cite{pitsilis} built an ensemble model of RNN classifiers for identifying hateful posts from a large dataset of Twitter posts. Studies are also being carried out on hate speech detection in languages other than English. Ibrohim \etal \cite{ibrohim} employed various classifiers like SVM, Random Forest Decision Tree, Naive Bayes for multi-label Indonesian tweets classification. Vo \etal \cite{Quan} employs a multi-channel CNN-LSTM network to detect hate speech in the Vietnamese language. 
\par
Although the majority of the approaches are based on monolingual content, some approaches are proposed for hate speech detection of code-switched texts. One of the earlier works for code switched texts was presented by \cite{bhatia} demonstrating cross-lingual interaction on the semantic level. There have been various attempts to translate the Hindi-English mixed language into pure English previously, but the major obstacle to this is that the grammatical rules of Hinglish are very uncertain and user-dependent. \cite{joshi} used sub-word level LSTM models for Hinglish sentiment analysis. \cite{choudhary} proposed the use of contrastive learning with Siamese networks to map code-mixed and standard language text to a common sentiment space. Hate speech can often be complex and hence \cite{lal} has used two kinds of encoders that take note of overall sentiment and individual sentiment-bearing units. In addition to this, they have also used a featured network that uses linguistic features to augment the model. 
\par
Baroi \etal \cite{baroi} uses CNNs and LSTMs based ensemble models to detect hate speech. CNN-based transfer learning has also been used for the detection of Hinglish hate speech \cite{mathur-2018-trans}. Gupta \cite{gupta} has utilized bi-directional sequence models such as GRU, BiLSTM, etc with data augmentation techniques such as synonym replacement, Random Insertion, Random Swap, Random Deletion on the text to achieve scores. However, the majority of the approaches have relied on recurrence or convolution along with complex data processing steps for offensive text classification in code-switched languages. Also, the comparatively recent attention mechanism has not yet been actively used in this domain. While Chopra \etal \cite{chopra} does make use of attention, they use complex text encoding steps for bias elimination and also combine the attention layer with LSTM in the final architecture. Further, advanced transformer architecture \cite{leite, kamal} have also been tried for code-switched hate speech detection but the higher complexity of these models makes them impractical to use in real-world.

\section{Proposed Methodology}
\par
Our task is to classify a given text sample into one of the three categories of non-abusive, abusive, and hate-inducing. We explain our solution by describing our approach for each phase of the standard text classification pipeline- preprocessing, text encoding, and model architecture.
\subsection{Data Preprocessing}
\par
The tweets obtained through data sources were passed through a pre-processing pipeline. The pre-processing pipeline can be broken down into intermediate steps as follows:

\begin{itemize}
    \item Lower case: The data is transformed to lowercase throughout.
    \item Replace emojis: Emojis that appear in tweets are replaced with relevant textual information with the help of ‘emoji’ an Emoji for python library
    \item Strip hashtags, user mentions, and HTML tags: Hashtags, user-mentions, and HTML links that are often used in tweets are removed. User-mentions are replaced with the keyword “username”. Links are replaced with the keyword “link” and hashtags are replaced with corresponding plain text. 
    \item Expand Contractions: The apostrophe is a punctuation mark that is often used to abbreviate a word or a group of words. For example, the word "don't" means "do not," while "can’t" means "can not". In this phase, the abbreviated forms are extended.
    \item Remove special characters: Special characters are neither alphabets nor numbers and these induce noise hence are removed from the text data.
    \item Transliterate: This step involves transliteration of the Hinglish text data i.e conversion of Hindi text into relevant sounding English text. This step is carried out using the indic\_transliteration package’s “sanscript” library.

\end{itemize}

\subsection{Text Encoding}
\par
We make use of subword tokenization to convert the text into model-friendly data. Subword tokenization breaks the sentence into chunks based on the word frequency. As Hindi + English code-mixed data contains non-conventional vocabulary, this approach helps to solve the issues faced by word-based tokenization (large vocabulary, large number of OOV tokens, and different meanings of very similar words) and character-based tokenization (very long sequences and less meaningful individual tokens). Subword tokenization deals with an infinite potential vocabulary through a finite list of known words. For example, we can make up the word “unfortunately” via “un” + “for” + “tun” + “ate” + “ly”. The common words like “for”, “ate” are tokenized as whole words, while rarer words are broken into smaller chunks. Various subword tokenization algorithms like Byte Pair Encoding (BPE), Probabilistic Subword Tokenization, and Unigram Subword Tokenization have been used by researchers in the past. For our use, we focus on the BPE \cite{bpe} and \cite{unigram} subword tokenization algorithms. We make use of both BPE and Unigram algorithms as BPE might build an ambiguous tokenized sequence sometimes and Unigram algorithm helps to tackle this shortcoming of BPE.

\subsubsection{Byte Pair Encoding (BPE)}
\par
    In BPE, frequently occurring subwords are merged finding the ideal balance between character and word level representation. This helps to manage large corpora and encoding of any rare words in the vocabulary with appropriate subword tokens without introducing any “unknown” tokens. The BPE operates as follows:
    \begin{itemize}
        \item Get the word count frequency.
        \item Get the frequency of character level counts.
        \item Merge the most common byte pairing and add this to the list of tokens.
        \item Recalculate the frequency count for each token.
        \item Rinse and repeat until the defined token limit or number of iterations is reached.
    \end{itemize}
\subsubsection{Unigram Subword Tokenization}
\par
   The Unigram language model is another algorithm for subword segmentation. One of the assumptions is all subword occurrences are independent and subword sequence is produced by the product of subword occurrence probabilities. Unlike BPE, unigram helps to overcome a problem that we have no way to predict which particular token is more likely to be the best one when encoding any new input text rather than choosing the best option. The Unigram operates as follows:
    \begin{itemize}
        \item Choose the seed subword token set and the most frequently occurring substrings. 
        \item Calculate the probability for each subword token.
        \item Calculate a loss value of each subword. The Expectation-Maximization (EM) algorithm is used to calculate the loss.
        \item Drop the bottom x\% of the subword tokens based on the loss. To avoid OOV words, single characters are kept.
        \item Rinse and repeat until the desired vocabulary size or there is no change in token numbers after successive iterations.
    \end{itemize}

\par
To improve the performance of the model, sequences are usually padded to a fixed length. This is usually performed by adding characters or truncating characters at the start (pre-padding) or at the end (post-padding). However in a task like hate speech detection, profanity is generally used by the user at the start or at the end of the sentence. Thus if we pad the sequences in only one particular way, there is a risk of losing information. To overcome this issue, we pass two inputs to our model for every sample. One sequence is obtained by pre-padding the BPE encoded sentence and the other by post-padding the sequence obtained by the Unigram algorithm. This approach of using pre-padded BPE and post-padded Unigram encoded sequence helps us to retain the features of the entire sequence and subsequently boost our results as shown later. It should be noted that the layers used in proposed model architecture are the same for both the sequences and parameters used for layers like positional encoding, and the multi-head attention layers are also the same. 

\subsection{Model Architecture}
\par
In this paper, we propose the use of a multi-headed self-attention mechanism \cite{attention} as the principal component of the model architecture. We make use of an attention mechanism to completely skip the recurrence mechanism from the model and reduce the memory requirement and complexity. Therefore, due to lack of recurrence mechanism, no information regarding the order of sequence is present. To overcome this problem, we use a positional encoding layer to provide information regarding the position of a word in the sequence. For each of the two inputs, the embedding layer output is passed to the positional encoding layer. Subsequently, the output from the positional encoding layer is passed to the multi-head attention layer. Further, the two representations obtained are concatenated together and then passed through a dense layer, a dropout layer, and then to the output nodes. The model architecture is as shown in fig 1. The working of each layer is as follows: 
\begin{figure}[]
\centerline{\includegraphics[width=5cm,height=10cm]{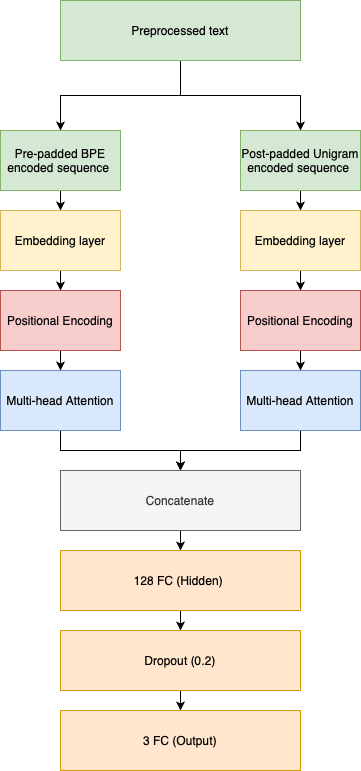}}
\caption{Block diagram of the proposed model architecture}
\label{fig 1}
\end{figure}
\subsubsection{Positional Encoding}
\par
The positional Encoding layer provides information regarding the absolute and relative position of tokens in the sequence. The layer connects properly as it has the same dimension as that of the embedding layer. In the past, both fixed as well as learned positional encodings have been employed by researchers. In our approach, we have used fixed positional encodings derived from sinusoidal functions of different frequencies:
\begin{equation}
    PE_{pos,2i} = sin(pos/10000^{2i/d_{model}})
\end{equation}
\begin{equation}
    PE_{pos,2i+1} = cos(pos/10000^{2i/d_{model}})
\end{equation}
\par
Where $pos$ is the position, $i$ is a particular dimension, and $d_{model}$ is the dimension of the embeddings. The advantage of these functions is that they are able to address relative positions properly. Every $n + k^{th}$ positional encoding could be represented as a linear function of $PE_n$. 

\subsubsection{Multi-head Attention}
\par
Attention function is the mapping of queries and key-value pairs to an output. In self-attention, different positions of a sequence are related to calculate representations of the same sequence. Based on correlation with other words present in the sequence, an attention vector is calculated to predict a new word. 
\par
Particularly in our case, we make use of scaled dot product variation of self-attention. The attention function calculates a dot product of query (Q) and key values (K) and the result of the dot product is passed through Softmax before obtaining final weights to be multiplied with the values (V). The formula for scaled dot product attention is as follows: 

\begin{equation}
    Attention (Q,K,V) = softmax(\dfrac{QK^T}{\sqrt{d_k}})V
\end{equation}

\hspace{-10pt}where, \\
V is the value vector, \\
Query (Q) = $EW_q$, \\
Key (K) = $EW_k$, \\
Value (V) = $EW_v$.
\par
$W_q$, $W_k$, $W_v$ are the respective weight matrices for queries, keys, and values. $1/\sqrt{d_k}$ is used as a scaling factor, and thus it is named as scaled dot product attention. In our case, $Q$, $K$, and $V$ are all the same representations obtained from the positional encoding layer. 
\par
As we make use of multi-head self-attention, attention function output for n different projections of the queries, keys, and values are calculated rather than making use of only one attention function output. Further, all the obtained output values are concatenated together and further processing takes place. The function of multi-head self-attention is given as follows: 

\begin{equation}
    Multihead(Q,K,V) = Concat(h_1, ..., h_n)W^o
\end{equation}
where, \\
$ h_i = Attention(Q(W_i)^Q, K(W_i)^K, V(W_i)^V)$
\par
Based on our empirical studies, we found n = 8 to be the ideal number of parallel attention layers for our task. To obtain the final value, the values of parallel attention layers are concatenated together as shown in fig 3. 
\par
After this stage, concatenation of the outputs obtained for two input sequences takes place which is then passed to the next layers in the architecture. 
\begin{figure}[]
\centerline{\includegraphics[width=5cm,height=5cm]{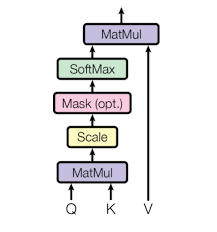}}
\caption{Schematic representation of self dot product attention \cite{attention}}
\label{fig 2}
\end{figure}
\begin{figure}[]
\centerline{\includegraphics[width=4cm,height=5cm]{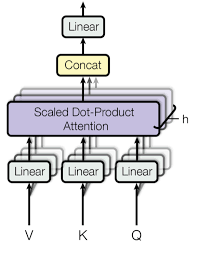}}
\caption{Multi head attention consisting of parallel running attention layers \cite{attention}}
\label{fig 3}
\end{figure}

\subsubsection{Dense and Dropout}
\par
After the concatenation of two outputs obtained from previous layers, the output is passed to a dense layer containing 128 nodes. In the dense layer, every neuron receives input from all the neurons in the previous layer i.e. they are deeply connected. The dense layer helps to change the dimension of the vector. It does so by performing a matrix-vector multiplication. The values in the matrix are parameters that are trained and updated through backpropagation. Further, to avoid overfitting, we use a dropout layer next. The dropout layer ignores a certain set of neurons at random while training the model so that no intra network co-dependency is formed \cite{dropout}. As per our experimentation, we found a dropout value of 0.2 to be ideal for the architecture. The output from the dropout layer is passed through to a dense layer containing three nodes that represent the final output layer, with each node corresponding to one label. Additionally, we train our model by optimizing the sparse categorical cross-entropy loss.

\section{Dataset Description and Results}
\subsection{Dataset Description}
\par
Two datasets have been used for the study and analysis in this paper which are developed by \cite{mathur-2018-trans} and \cite{davidson}. Table 1 shows the tweet distribution in the English dataset provided by \cite{davidson} and the HEOT dataset provided by \cite{mathur-2018-trans}. The HEOT dataset consisting of 3679 tweets was developed using the Twitter Streaming API by gathering specified profane terms in Hinglish language and choosing tweets in Hindi-English code-switched language. Another corresponding labeled dataset for English tweets consists of 14509 records. This dataset was collected using the Twitter API which collected samples from 33485 users, this resulted in a collection of 85.4 million tweets from which random samples were labeled manually. 

\begin{table}[htbp]
\centering
\caption{Tweet distribution in Davidson and HEOT dataset.}
\begin{tabular}{|c|c|c|}
\hline
Label         & HEOT & Davidson \\ \hline
Non-abusive   & 1414 & 7274     \\ \hline
Abusive       & 1942 & 4836     \\ \hline
Hate-inducing & 323  & 2399     \\ \hline
Total         & 3679 & 14509    \\ \hline
\end{tabular}
\end{table}

\par
It can be observed that the size of the HEOT dataset is noticeably small as compared to the English tweets dataset. In reality, users identifying to a specific demographic division are small in comparison to the total users. Thus, this unusual distribution is advantageous as the small size of Hinglish tweet samples represents a true world scenario. Further, the tweets are classified into three categories: non-offensive, abusive, and hate-inducing. Example of each category of the HEOT dataset and their English translation are given in Table 2. 

\begin{table}[htbp]
\centering
\caption{Label-wise HEOT dataset example and their English translation}
\begin{tabular}{|c|c|c|}
\hline
Label         & HEOT & English \\ \hline
Non- & RT @username & RT @username I \\
abusive &  HNY k time pe  & had this earned \\
 & 15000 kamaya  & 15000 during \\
 & tha maine.. & new year.. time \\
 & is baar payment & people are \\
 & nahi de rahe hai & not paying  \\ \hline
Abusive       & @username K*tiya!& @username B*tch! \\
& Mujhe mat sikha:/ & Do not teach me  \\ \hline
Hate- & @username Gujraat   & @username People\\
inducing &wale Teri tarah  & from Gujrat are not \\ 
&chu*iya nhi ...&f*ckers like you... \\
&Nation first...&Nation first ... \\
&f*ck all Muslims&f*ck all Muslims\\ \hline
\end{tabular}
\end{table}

\subsection{Results}
\par
The results obtained on the test set are evaluated across accuracy, weighted precision, weighted recall, and weighted F1 score. We compare the proposed model with approaches used by researchers in the past and different combinations of padding types used in the ensemble method. The results are as shown in Table 3. 
\par
It can be seen that the attention ensemble model using pre-padded BPE encoded sequence and post-padded Unigram encoded sequence outperforms individual models as well as ensemble models with different padding structures. Next, we can also see that the proposed ensemble model has performed better than the models employed by researchers in the past. 

\par
Next, we show the impact of BPE and Unigram text encoding algorithms by combining attention with 128 units of BiLSTM and using BPE and Unigram text encoded sequences in Table 4. We can see that with BiLSTM layers, the performance obtained is better, but it comes at the cost of increased complexity. Further,  we also carry out a complexity analysis of various layer types as shown in Table 5. Thus, attention tends to be less complex as compared to conventional approaches.

\par
Therefore, a brief result analysis shows the benefit of using a multi-head self-attention mechanism over conventional approaches with slight variation in performance based on changes in certain factors. The use of attention and recurrent components yields slightly better performance than the proposed approach, but its complexity is substantially higher than the proposed approach. 

\begin{table}[htbp]
\centering
\caption{Performance metrics of various models. Post and Pre refers to post-padding and pre-padding.
}
\begin{tabular}{|c|c|c|c|c|}
\hline
Model                                   & Accuracy       & F1             & Precision      & Recall         \\ \hline
Attention + BPE (Post)  & 87.02          & 0.842          & 0.848          & 0.865          \\
+ Unigram (Post) & & & &\\
\hline
Attention + BPE (Pre)      & 87.04          & 0.846          & 0.851          & 0.864          \\ 
+ Unigram (Pre) & & & &\\\hline
Attention + BPE (Pre)    & \textbf{87.41} & \textbf{0.851} & \textbf{0.862} & \textbf{0.868} \\ 
 + Unigram (Post) & & & &\\\hline
Attention + BPE (Post)    & 86.17          & 0.836          & 0.842          & 0.853          \\ 
+ Unigram (Pre)  & & & &\\\hline
Attention + BPE (Post)                    & 86.15          & 0.833          & 0.839          & 0.854          \\ \hline
Attention + Unigram (Post)                & 86.38          & 0.839          & 0.844          & 0.859          \\ \hline
Attention + BPE (Pre)                     & 86.06          & 0.829          & 0.834          & 0.848          \\ \hline
Attention + Unigram (Pre)                 & 86.19          & 0.838          & 0.846          & 0.855          \\ \hline
Joshi \etal \cite{joshi}                  & 69.7           & 0.658          &  NA              &  NA              \\ \hline
Choudhary \etal \cite{choudhary}                      & 77.3           & 0.759          & 0.770          & 0.749          \\ \hline
Mathur \etal \cite{mathur-2018-trans}                      & 83.90          & 0.714          & 0.802          & 0.698          \\ \hline
Gupta \cite{gupta}                          & 79             & 0.706          & 0.733          & 0.693          \\ \hline
Lal \etal \cite{lal}                              & 83.54          & 0.827          & NA               & NA               \\ \hline
Chopra \etal \cite{chopra}                                 & 85             & 77             & NA               & NA               \\ \hline
\end{tabular}

\end{table}

\begin{table}[htbp]
\centering
\caption{Comparison of results for combination of Attention and LSTM with BPE and Unigram encoding}
\begin{tabular}{|c|c|c|}
\hline
Model                            & Accuracy & F1    \\ \hline
Attention + LSTM + BPE           & 88.14    & 0.875 \\ \hline
Attention + LSTM + Unigram       & 88.33    & 0.877 \\ \hline
Attention + LSTM + BPE + Unigram & 88.56    & 0.878 \\ \hline
\end{tabular}

\end{table}

\section{Analysis}
Our main observations after performing out the study have been
\begin{itemize}
    \item Recurrent models perform slightly better than the self attention-based models, but the minimum number of sequential operations required and complexity per layer of self-attention is less in comparison to the recurrent or convolution models. Therefore, the use of self-attention is a desirable choice.
    \item Use of BPE and Unigram encoded sequences with attention leads to impressive results, especially in the case of Hindi-English code-switched data where the vocabulary is full of non-conventional words. 
    \item Concatenation of pre-padded BPE and post-padded Unigram encoded sequences leads to substantial improvement in results. 
    \item Recent development of models based on transformer architecture like XLNet \cite{xlnet} tend to perform better than the proposed architecture, but they require huge computational power, and the training cost is also very high. In addition, these transformer-based architectures have multi-head self-attention as their building block only. 

\end{itemize}
\begin{table}[htbp]
\centering
\caption{Complexity analysis of various layers. n refers to the sequence length, d is the embedding dimension, and k refers to the kernel size}
\begin{tabular}{|c|c|}
\hline
Layer type  & Complexity per layer \\ \hline
Attention   & $O(n^2.d)$              \\ \hline
Recurrent   & $O(n.d^2)$              \\ \hline
Convolution & $O(k.n.d^2)$            \\ \hline
\end{tabular}

\end{table}

\section{Conclusion}
\par
We have been able to achieve standard results in the field of Hindi-English code-switched language hate speech detection by making use of the attention mechanism. To overcome the shortcoming of attention that it does not retain all information of a sequence, it can be coupled with features like positional encoding while cutting down the memory requirement as well as complexity. While some earlier approaches have used very complex preprocessing steps to handle the non-conventional Hinglish vocabulary, our results show that by using simpler but efficient preprocessing steps like the use of BPE and Unigram subword tokenization algorithms we obtain better results. The improved results across metrics over the previous approaches are a testament to the power which the subword tokenization algorithms and attention mechanism hold. In the future, modifications in the post attention layers can be carried out for better information transfer. Also, the approach can be tried out for hate speech detection in other code-switched languages as they play a major role in the online structuring of multi-linguistic societies. With the dominance of models based on attention like Transformer, BERT, XLNet on traditional methods, the idea of using recurrence for sequence modeling is becoming weaker gradually. Models based on attention have proven to be a better alternative and can be used in sentiment classification effectively.

\bibliographystyle{IEEEtran}
\bibliography{references}
% \begin{thebibliography}{00}
% \bibitem{bhatia} Bhatia, Tej K., and William C. Ritchie. “The Bilingual Mind and Linguistic Creativity.” Journal of Creative Communications, vol. 3, no. 1, Mar. 2008, pp. 5–21, doi:10.1177/097325860800300102.

% \bibitem{joshi} Prabhu, Ameya et al. “Towards Sub-Word Level Compositions for Sentiment Analysis of Hindi-English Code Mixed Text.” COLING (2016).
% \bibitem{choudhary} Choudhary, Nurendra et al. “Sentiment Analysis of Code-Mixed Languages leveraging Resource Rich Languages.” ArXiv abs/1804.00806 (2018)
% \bibitem{b4} K. Elissa, ``Title of paper if known,'' unpublished.
% \bibitem{b5} R. Nicole, ``Title of paper with only first word capitalized,'' J. Name Stand. Abbrev., in press.
% \bibitem{b6} Y. Yorozu, M. Hirano, K. Oka, and Y. Tagawa, ``Electron spectroscopy studies on magneto-optical media and plastic substrate interface,'' IEEE Transl. J. Magn. Japan, vol. 2, pp. 740--741, August 1987 [Digests 9th Annual Conf. Magnetics Japan, p. 301, 1982].
% \bibitem{b7} M. Young, The Technical Writer's Handbook. Mill Valley, CA: University Science, 1989.
% \end{thebibliography}

\end{document}